\RequirePackage{fix-cm}
\documentclass[twocolumn]{svjour3}
\smartqed

\makeatletter
\let\cl@chapter\undefined
\makeatletter

\usepackage[pagebackref,colorlinks,linkcolor=red,anchorcolor=blue,citecolor=green,CJKbookmarks=True]{hyperref}
\usepackage{amsmath,amssymb,amsfonts}
\usepackage{url}
\usepackage{booktabs}
\usepackage{graphicx}
\usepackage{natbib}
\usepackage{bbding}
\usepackage{multirow}
\usepackage{booktabs}
\usepackage{dsfont}
\usepackage{times}

\usepackage{mathrsfs}%
\usepackage[title]{appendix}%
\usepackage{xcolor}%
\usepackage{textcomp}%
\usepackage{manyfoot}%
\usepackage{algorithm}%
\usepackage{algorithmicx}%
\usepackage{algpseudocode}%
\usepackage{listings}%
\usepackage{array}%
\usepackage{float}%
\usepackage{bm}%

\begin{document}

\title{LaMD: Latent Motion Diffusion for Image-Conditional Video Generation
}

\author{Yaosi Hu \and
        Zhenzhong Chen ~\textmd{\Envelope} \and
        Chong Luo
}

\institute{
Yaosi Hu \and Zhenzhong Chen (Corresponding author) \at
School of Remote Sensing and Information Engineering, Wuhan University, Wuhan, China. \\ 
\email{ys\_hu@whu.edu.cn, zzchen@whu.edu.cn} \\
\and
Chong Luo \at
Microsoft Research Asia, Beijing, China. \\
\email{cluo@microsoft.com}
}

\date{Received: date / Accepted: date}

\maketitle

\begin{abstract}
    The video generation field has witnessed rapid improvements with the introduction of recent diffusion models. While these models have successfully enhanced appearance quality, they still face challenges in generating coherent and natural movements while efficiently sampling videos. In this paper, we propose to condense video generation into a problem of motion generation, to improve the expressiveness of motion and make video generation more manageable. This can be achieved by breaking down the video generation process into latent motion generation and video reconstruction. 
    Specifically, we present a latent motion diffusion (LaMD) framework, which consists of a motion-decomposed video autoencoder and a diffusion-based motion generator, to implement this idea.
    Through careful design, the motion-decomposed video autoencoder can compress patterns in movement into a concise latent motion representation. Consequently, the diffusion-based motion generator is able to efficiently generate realistic motion on a continuous latent space under multi-modal conditions, at a cost that is similar to that of image diffusion models. Results show that LaMD generates high-quality videos on various benchmark datasets, including BAIR, Landscape, NATOPS, MUG and CATER-GEN,  that encompass a variety of stochastic dynamics and highly controllable movements on multiple image-conditional video generation tasks, while significantly decreases sampling time.

\end{abstract}

\begin{sloppypar}

\section{Introduction}\label{sec:intro}

\begin{figure*}[htbp]
    \centering
    \includegraphics[width=1.0\textwidth]{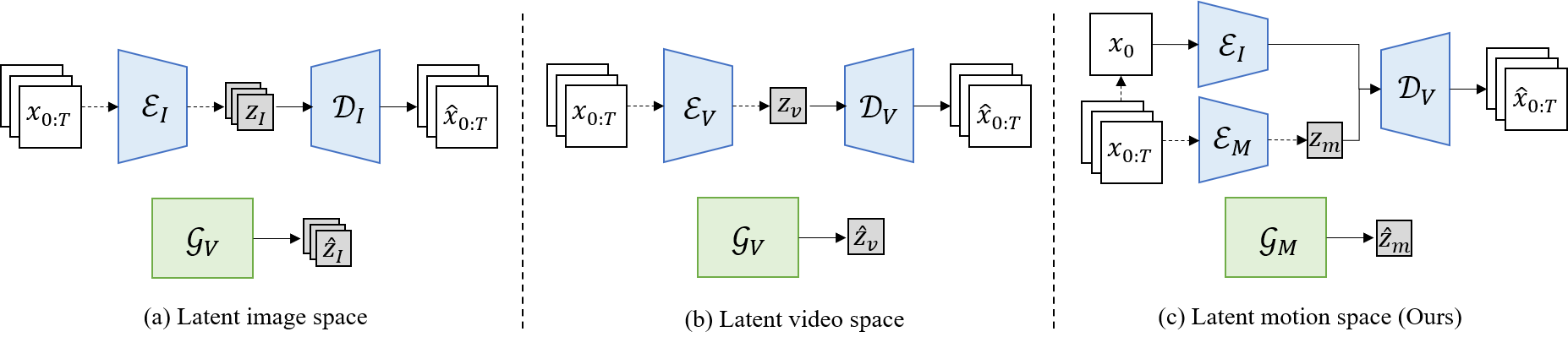}
    \caption{The comparison of video generation in different latent space. The dashed line stands for operations only involved in training process, while the solid line represents operations both involved in training and sampling process.}
    \label{fig:paradigm}
\end{figure*}

Video generation aims to generate natural, high-quality videos that accurately reflect human intentions. Despite ongoing efforts, the field has yet to fully achieve this goal. One existing and straightforward solution to the video generation problem is to train a deep model which directly generates the pixel values for each frame in a video \citep{10.1007/978-3-030-01240-3_37, 9577375, ho2022video}. However, the sheer volume of video data makes it challenging to train the model effectively, and this lack of effectiveness has two implications. First, the training procedure requires significant computational resources, which can hinder the scalability of the model. Second, it may cause the generator to focus too much on spatial high-frequency details that are less noticeable to the human visual system. Therefore, developing a novel video generation framework that can efficiently and effectively generate videos with a focus on perceptually relevant information is essential. 

As an alternative to the above pixel-space generation approach, latent-space generation has become a preferred choice for reducing data redundancy and conserving resources during model training. This is achieved by using a pre-trained autoencoder to transfer the generation process from the pixel domain to a more efficient and compressed latent domain. Current latent-space video generation follows two typical paradigms, depending on the corresponding latent space. The first paradigm, as depicted in Fig.\ref{fig:paradigm}(a), employs an image autoencoder to transform individual frames into latent tokens, thereby generating the sequence of latent tokens in the latent image space \citep{han2022show}. The second paradigm, as depicted in Fig.\ref{fig:paradigm}(b), employs a 3D video autoencoder to translate video clips into a latent space, and then generates content within this latent video space \citep{yan2021videogpt}.

These two paradigms for latent-space video generation each have their own strengths and weaknesses. The first paradigm leverages recent advancements in latent-space image generation \citep{ramesh2021zero, esser2021taming, rombach2022high} and is capable of producing high-quality, semantically accurate frames. However, since 2D image autoencoders do not consider the temporal relationships between frames, the generated collection of frames may be temporally incoherent and not accurately depict motion. Conversely, the second paradigm implements spatio-temporal compression from video clips to the latent video domain, thereby overcoming the issue of motion coherence \citep{he2022latent,zhou2022magicvideo,shen2023difftalk}. Nevertheless, modeling the video latent space, which requires 3D spatio-temporal information, presents significant challenges. 

To address this challenge, it is crucial to focus on modeling the most perceptually relevant information in the video. We have observed that the inferior performance in video generation compared to image generation is due to the difficulty in producing consistent and natural motion in videos. By separating the visual appearance and motion components of a video, we can concentrate on motion generation specifically. This idea underpins our design for a new paradigm to address the generation problem in latent motion space, as depicted in Fig.\ref{fig:paradigm}(c), which first separates motion from appearance and then applies generative model on compact latent motion space. This paradigm effectively preserves high-frequency details in image-conditional video generation tasks where appearance has already been provided. Additionally, it enhances the expressiveness of motion and reduces the complexity of modeling. Notably, the highly compressed motion representation enables the elimination of the 3D-CNN structure in generative models, resulting in a significant decrease in sampling time.

To this end, we present the Latent Motion Diffusion (LaMD) framework, which represents a practical implementation of the Latent Motion Generation paradigm. The LaMD framework consists of two components, namely the Motion-Content Decomposed Video Autoencoder (MCD-VAE) and a diffusion-based motion generator (DMG). The MCD-VAE component tackles the challenge of separating motion from appearance or content by utilizing a multi-scale image encoder, a lightweight motion encoder with an information bottleneck, and a decoder for video reconstruction. The DMG module is influenced by the highly effective generative performance of Diffusion Models (DMs) \citep{ho2020denoising, dhariwal2021diffusion}. It gradually denoises motion variables over a continuous latent space and can also incorporate pre-obtained content features, as well as optional text, as multi-modal conditions. The computational complexity of the DMG module is comparable to that of image diffusion models, resulting in reduced resource utilization and faster sampling speed compared to other video diffusion models.

Our proposed LaMD framework has been evaluated on multiple image-conditional video generation tasks using five commonly used datasets. The results show that the MCD-VAE achieves high reconstruction quality, with excellent compression of motion, and the DMG generates natural motion for video. LaMD also achieves competitive performance on a variety of benchmarks, including BAIR, Landscape, NATOPS, MUG and CATER-GENs which cover a wide range of motions from stochastic dynamics to highly controllable movements. In conclusion, our work makes the following contributions:
\begin{itemize}
    \item  We have introduced Latent Motion Generation as a new paradigm for video generation. The new paradigm focuses on the expressiveness of motion and therefore has the potential to generate videos with high perceptive quality. 
    \item We have established a practical implementation of this paradigm in the form of the Latent Motion Diffusion (LaMD) framework. The LaMD framework, which is made up of MCD-VAE and DMG, can produce videos with coherent and realistic motion under the appearance constraint. The option to incorporate text as a condition further increases control over motion.
    \item We have demonstrated the high-quality video generation capability of the LaMD framework through its results on five benchmark datasets for stochastic image-to-video (I2V) generation, class-guided image-to-video (cI2V) generation and text-image-to-video (TI2V) generation tasks. Notably, we have also achieved a substantial decrease in sampling speed, comparable to image diffusion models.
\end{itemize}

The rest of the paper is organized as follows: Section \ref{sec.related} provides a brief review of the related work on various paradigms for video generation. Section \ref{sec.method} presents a detailed explanation of the proposed method, while Section \ref{sec.exp} showcases and analyzes the experimental results thoroughly. Finally, we conclude this paper and discuss its limitations and future prospects in Section \ref{sec:discuss}.

\section{Related Work}\label{sec.related}

\textbf{Video Generative Models} have been developed in various types, including Variational Autoencoders (VAEs) \citep{kingma2014auto}, Generative Adversarial Networks (GANs) \citep{goodfellow2020generative}, autoregressive models (ARMs) \citep{chen2020generative}, and Diffusion Models (DMs) \citep{ho2020denoising}. These models have demonstrated remarkable success in generating high-quality videos. VAEs attempt to learn latent variables $z$ close to a prior distribution $P_{z}$ by parameterizing the data distribution with a surrogate loss, but they may suffer from posterior collapse. While VAEs enable efficient sampling, the synthesized results are often blurrier than those of GANs, which are likelihood-free and based on the contest between two networks. GANs can achieve high-resolution synthesis with good perceptual quality but are unstable for training. 
Recently, DMs have shown promising results in various tasks which gradually inject and remove noise from data, but their sampling speed is much slower, as is that of ARMs. These generative models are earlier applied to generate videos in pixel domain. Recently, some work prefer to use a powerful autoencoder to transfer the generative goal from pixel space to latent space, which alleviates the burden of generative models and improves the sample quality.

\textbf{Pixel Space Video Generation} employs generative models to generate videos at the pixel level. Most of the existing methods focus on modeling the global motion representation from videos, i.e. by approaching a prior normal distribution to enable stochastic generation of videos from randomly sampled motion during inference \citep{babaeizadeh2018stochastic, tulyakov2018mocogan, zhang2020dtvnet, wang2020g3an, 9577375, saito2020train}. Some methods incorporate additional clues on structure or action to support controllable generation \citep{sheng2020high, menapace2021playable, blattmann2021understanding, xu2023conditional}. In Addition, DMs are recently used to model video space \citep{ho2022video}. However, due to the high computational complexity of video data, these approaches are challenging for high-resolution video generation unless a spatial-temporal super-resolution module is further added to generative models \citep{singer2022make, ho2022imagen}. Moreover, image compression techniques have shown that in general more than 80\% of the data could be removed without noticeable decrease of perceptual quality \citep{wallace1991jpeg}. But pixel-level generative models spend most of capacity to generate these imperceptible information, resulting in inefficient training. As a result, more recent researches focus on latent space video generation approaches.

\textbf{Latent Space Video Generation} typically adopts an autoencoder to handle the compression and reconstruction between pixel space and latent space, with subsequent generative model focused solely on latent space. Common image autoencoders like VQ-VAES \citep{NIPS2017_7a98af17, 10.5555/3454287.3455618} and VQ-GANs \citep{esser2021taming, yu2022vectorquantized} achieve high-quality synthesis which compress image into discretized latent tokens. \citep{zheng2022movq} has further modified it by modulating the quantized vectors to integrate spatially variant information, while \citep{rombach2022high} replaces the quantization layer with a slight KL-penalty over continuous latent variables. 
These powerful image autoencoders are widely employed in image generation models and help to achieve impressive performance on generating high quality and realistic images \citep{pmlr-v139-ramesh21a, Lee_2022_CVPR, chang2022maskgit, Ramesh2022HierarchicalTI}. Building on advancements in image generation, some works utilize image autoencoders to create various video generative models that function in latent image space \citep{walker2021predicting, shrivastava2021diverse, seo2022harp, wu2022nuwa, hong2022cogvideo, Han_2022_CVPR}. Meanwhile, other prior research extends a pre-trained image diffusion model into video generators by integrating additional temporal modules. \citep{Han_2023_ICCV, Blattmann_2023_CVPR, 10.1145/3581783.3612707, guo2024animatediff}. 
Since using image autoencoder to encode video into frame-wise latents can break the pixel-dependency across frames, \citep{yan2021videogpt, villegas2022phenaki,he2022latent,10.1007/978-3-031-19790-1_7,Fu_2023_CVPR,10.1007/978-3-031-72986-7_23} extend encoder from 2D-based to 3D-based, supporting video generation on latent video space. However, generating latent content and motion information simultaneously can be a challenging goal which would result in inefficient training.
Following the generation paradigm within the latent image/video space, some concurrent work focus on image-conditional video generation to incorporate image as instruction which show significant success in generating motion-rich videos \citep{chen2023videocrafter1, blattmann2023stable, zhang2023i2vgen, xing2023dynamicrafter, Ni_2024_CVPR, gal2024breathing, Hu_2024_CVPR, Zeng_2024_CVPR}. These models typically leverage extensive video datasets for training or are enhanced by off-the-shelf text-to-video generation models. Despite their ability to produce high-quality videos, they tend to require considerable model complexity to learn latent video representation and exhibit slow sampling speeds. Our work introduces an innovative paradigm of latent motion generation that offers substantial potential for generating high-quality videos while markedly diminishing the learning burden on the model.

The evolution of video generation from pixel space to latent space aims at shifting the function of latent-to-pixel synthesis to a pre-trained autoencoder, allowing the generative model to concentrate on generating latent video instead. In this paper, we take it a step further by transferring the motion-content synthesis to the autoencoder, enabling generative models to focus solely on motion generation. 
The most closely related work to ours is LFDM \citep{Ni_2023_CVPR}, which separates motion as a sequence of latent flows between frames and uses a 3D-UNet-based diffusion model to generate these sequences. Another similar approach \citep{10.1145/3641519.3657497} explicitly predicts optical flows for the initial image using a 3D-UNet-based diffusion model, which is then used to render the video with an additional diffusion model. In contrast, our approach provides a more compact and universal latent motion representation requiring only a 2D-UNet-based diffusion model, which enables a substantial increase in motion expressiveness and sampling speed.

\begin{figure*}[htbp]
    \centering
    \includegraphics[width=0.95\textwidth]{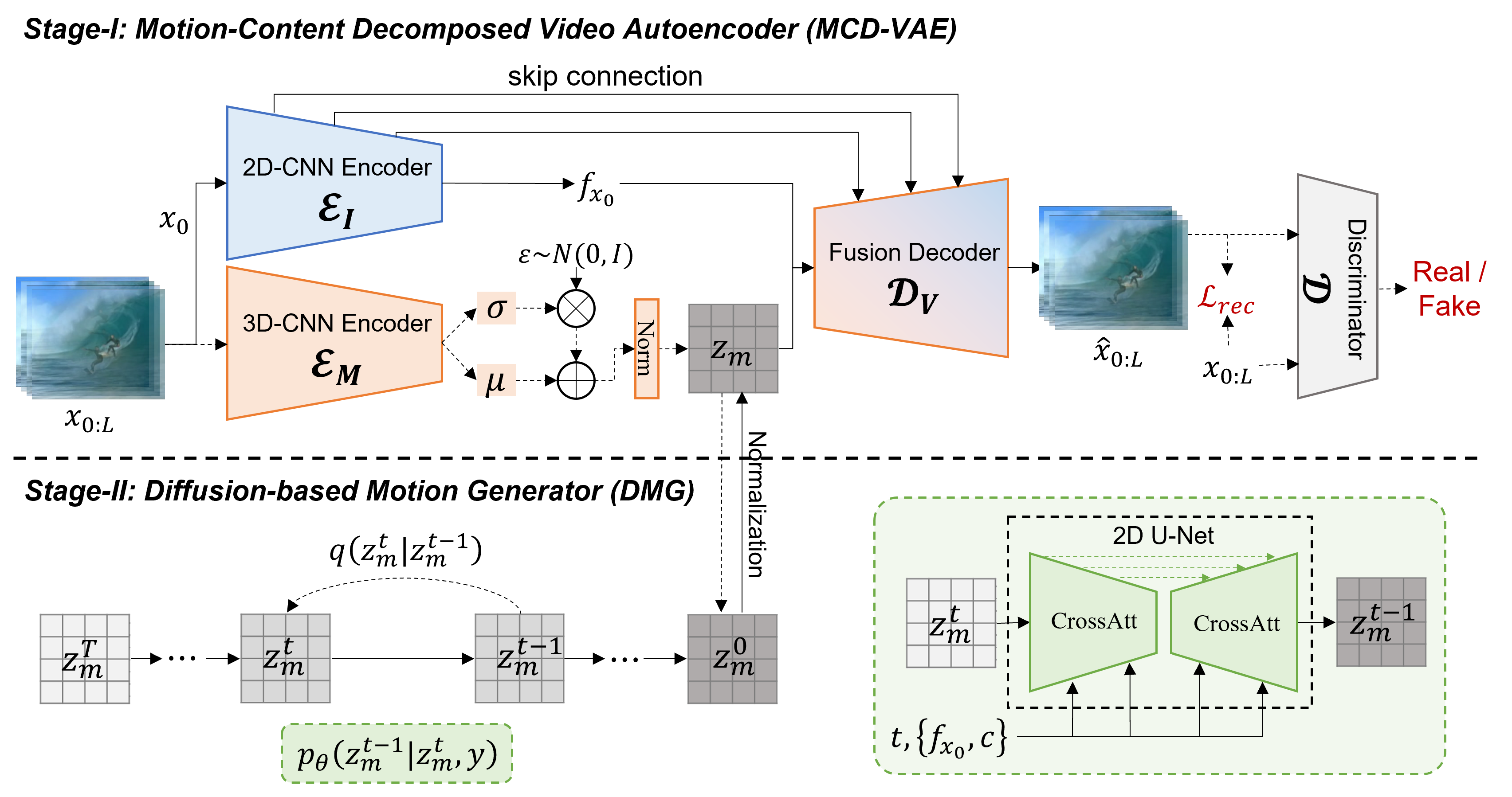}
    \caption{The framework of our proposed LaMD. During training process, the stage-I MCD-VAE is first trained to decompose latent motion with video reconstruction task, while DMG is trained to generate natural motion conditioned by $\left\{f_{x_0}, c\right\}$ in the second stage. During sampling process, the motion latents are first generated by DMG and then input into the decoder $\mathcal{D}_V$ together with multi-scale content features from the first given image to synthesize videos. The black dashed lines stand for operations only involved in training process.}
    \label{fig:framework}
\end{figure*}

\section{Method}\label{sec.method}
The essence of motion lies in the dynamic movements across frames, such as trajectories. These movements are highly redundant in both spatial and temporal dimensions and can be naturally separated from content (i.e., appearance). Thus, to support video generation in latent motion space, we first design a motion-content decomposed video autoencoder (MCD-VAE) which separates the representation of content and motion and further fuses them to reconstruct videos. Then a proposed diffusion-based motion generator (DMG) targets to generate content-specific motion latents conditioned on the pre-obtained content features with other optional conditions like text.

Our proposed two-stage approach LaMD is illustrated as Fig.\ref{fig:framework}. During training, we first train a MCD-VAE in a self-supervised manner using video-only data, and then fix its parameters and train the DMG in the latent motion space. During sampling, the DMG generates motion latents progressively from the normal distribution, guided by content feature extracted from the given image and other optional conditions. These motion latents are then fed into the MCD-VAE decoder to synthesize a video at once.

\subsection{Motion-Content Decomposed Video Autoencoder}
Targeting image-conditional video generation, preserving more content information is beneficial to improve the video quality, while the motion component can be significantly compressed due to its high redundancy on both spatial and temporal dimension. Thus, we employ different compression strategies through two separate encoders to extract motion and content representation.

\subsubsection{Decomposition of motion and content}
The essence of our decomposition strategy is creating one branch that only allows content information to pass, thereby ensuring that the remained motion information is restricted to the other branch. By controlling the capacity of the second branch, we are able to extract a decoupled motion representation that avoids the content leak while achieving perfect video reconstruction.

Following this design, our MCD-VAE consists of the 2D-CNN based image encoder $\mathcal{E}_I$, the 3D-CNN based motion encoder $\mathcal{E}_M$, the fusion decoder $\mathcal{D}_V$.
Given a video $x_{0:L} \in \mathbb{R}^{L \times H \times W \times 3}$ that contains $L$ frames, the image encoder $\mathcal{E}_I$ based on 2D-UNet architecture \citep{ronneberger2015u} first encodes the first frame $x_{0}$ into content latents $f_{x_{0}}\in \mathbb{R}^{h \times w \times d'}$, as well as intermediate multi-scale representations $f_{x_0}^1, \cdots, f_{x_0}^k$ through $\left\{f_{x_0}, f_{x_0}^1, \cdots, f_{x_0}^k\right\}=\mathcal{E}_I\left(x_0\right)$, where $k$ represents the number of scales. 
Meanwhile, the motion encoder $\mathcal{E}_M$ extracts the motion latents from video based on a light-weight 3D-UNet architecture \citep{cciccek20163d} with respective spatial and temporal downsampling factor $r_s, r_t$. 
The obtained latent motion $z_m$ is then passed through a normalization layer, which we have found to effectively improve the stability and efficiency of training the diffusion-based generative model. 
Additionally, it is worth noting that we encode the motion information into the channels by setting $r_t=L$, thereby eliminating the temporal dimension. This allows for the transfer of motion information from the temporal dimension into the channel dimension in the latent space without compromising reconstruction performance, but significantly streamlines the generation model with a more lightweight 2D-based architecture. Therefore, the latent motion representation is quite compact and of low dimensionality with $z_m \in \mathbb{R}^{h \times w \times d}$ where $h=\frac{H}{r_s},w=\frac{W}{r_s}$, and $d$ is the channel size.

Specifically, to achieve the decomposition of motion representation, an additional constraint is required on the motion branch. 
Here we create an information bottleneck \citep{alemi2017deep}, an effective technique to minimize the unnecessary information, on the top of motion encoder. This information bottleneck can be implemented using the reparameterization trick as
\begin{equation}
\mathbf{z}_m=\mu_\theta\left(\mathcal{E}_M\left(x_{0: L}\right)\right)+\mathbf{\varepsilon} \cdot \sigma_\theta\left(\mathcal{E}_M\left(x_{0: L}\right)\right)
\end{equation}
where $\bm{\varepsilon} \sim N(\mathbf{0}, \mathbf{I})$. 
Accordingly, we enforce a KL-penalty constraint on the distribution of latent motion controlling the capacity of motion branch with an adjustable hyper-parameter $\beta$ (the KL term in Eq.\ref{equ:loss_vae}), to force the latent motion $z_m$ as a minimal sufficient statistic of temporal information to reconstruct video. Since the image branch already provides the required content information for reconstruction, reducing the bottleneck size results in the exclusion of content information from motion branch. Therefore, by adjusting the strength of the KL-penalty constraint, we can control the capacity of motion branch to squeeze the appearance out of motion branch and maintain the remained motion information, thereby achieving motion and content decomposition. The decomposition ability can be validated through the experiment in Sec.\ref{sec:exp_vae}.

\subsubsection{Video reconstruction}
\begin{figure}[!t]
    \centering
    \includegraphics[width=0.45\textwidth]{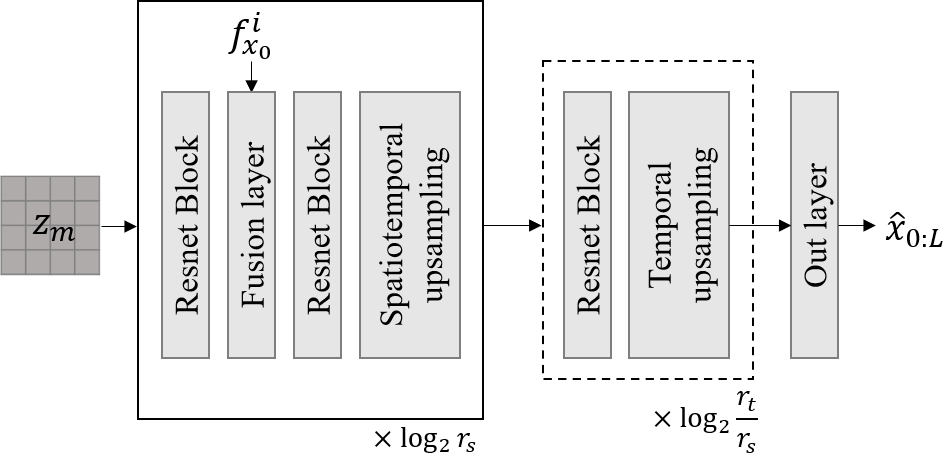}
    \caption{The architecture of fusion decoder}
    \label{fig:fusion_decoder}
\end{figure}

The fusion decoder $\mathcal{D}_V$ aims at fusing content and motion to reconstruct video pixels, with the architecture illustrated in Fig.\ref{fig:fusion_decoder}. Building upon the 3D-UNet decoder, we first add a fusion layer between the Restnet blocks to merge the input feature with the corresponding-scale content feature. This fusion is achieved through a straightforward concatenation operation followed by a 1$\times$1 convolution layer, which serves to normalize the dimensions of the video feature. Once all content integration blocks have been applied, the spatial resolution of the visual feature is restored. However, due to the elimination of the temporal dimension of latent motion $z_m$ in the motion encoder, which means $r_t=L>r_s$, we incorporate extra temporal interpolation blocks to align with the temporal dimension of the original video. Ultimately, the reconstructed video $\hat{x}_{0: L}$ is output after an output convolution layer.

The objective loss function contains the reconstruction loss which combines the $L_{1}$ pixel-level distance and LPIPS perceptual similarity loss \citep{10.5555/3157096.3157170}, and the KL divergence with hyper-parameter $\beta$ to control a clean decomposition of motion latents. Besides, based on \citep{esser2021taming}, we also apply an adversarial objective via a video discriminator $\mathcal{D}$ to improve the realism of reconstructions. Thus, the overall objective is formulized as 
\begin{equation}
\label{equ:loss_vae}
    \begin{aligned} 
    \underset{\mathcal{E}_I, \mathcal{E}_M, \mathcal{D}_V}{\operatorname{argmin}} &\max _{\mathcal{D}} \mathbb{E}_{x \sim p(x)}\left[\mathcal{L}_{G E N}+\lambda \mathcal{L}_{G A N}\right], \\
    \mathcal{L}_{G A N}= & \log \mathcal{D}\left(x_{0: L}\right)+\log \left(1-\mathcal{D}\left(\hat{x}_{0: L}\right)\right), \\
    \mathcal{L}_{G E N}=&\left\|x_{0: L}-\hat{x}_{0: L}\right\|_1+\operatorname{LPIPS}\left(x_{0: L}, \hat{x}_{0: L}\right) \\ &+\beta \operatorname{KL}\left(q_{\mu_\theta,\sigma_\theta,\mathcal{E}_M}\left(\mathbf{z}_m \mid x_{0: L}\right) \| N(\mathbf{0}, \mathbf{I})\right), 
    \end{aligned}
\end{equation}
where $\lambda$ stands for the adaptive weight applied in \citep{esser2021taming}.

\subsection{Diffusion-based Motion Generator}
Based on the normalized motion on continuous latent space, we apply a diffusion-based motion generator (DMG) to learn the motion distribution $p(\mathbf{z}_m)$ via the reverse process of a fixed Markov Chain. 

Our DMG is built on denoising diffusion probabilistic model (DDPM) \citep{sohl2015deep,ho2020denoising} that consists of a forward diffusion process and a reverse denoising process. Given a latent motion $\mathbf{z}_m^0 \sim p(\mathbf{z}_m)$ obtained from MCD-VAE, the forward process is gradually adding Gaussian noises to $\mathbf{z}_m^0$ according to a series of scheduled variances $\beta_1, \cdots, \beta_T$, producing a series of motion with increasing levels of noise $\mathbf{z}_m^1, \cdots, \mathbf{z}_m^T$ as
\begin{equation}
    q\left(\mathbf{z}_m^t \mid \mathbf{z}_m^{t-1}\right)=\mathcal{N}\left(\mathbf{z}_m^t; \sqrt{1-\beta_t} \mathbf{z}_m^{t-1},\beta_t \mathbf{I}\right).
\end{equation}
In practice, we can directly sample $\mathbf{z}_m^t$ at any time step from the marginal probability distribution $q\left(\mathbf{z}_m^T \mid \mathbf{z}_m^0\right)$:
\begin{equation}
    q\left(\mathbf{z}_m^t \mid \mathbf{z}_m^0\right)=\mathcal{N}\left(\mathbf{z}_m^t; \sqrt{\bar{\alpha}_t} \mathbf{z}_m^0,\left(1-\bar{\alpha}_t\right) \mathbf{I}\right),
\end{equation}
where $\bar{\alpha}_t=\prod_{s=1}^t \alpha_s, \alpha_t=1-\beta_t$. When the $T$ is larger enough, the noised latent motion $\mathbf{z}_m^t$ will approximates a standard Gaussian distribution.

To recover $\mathbf{z}_m^0$, the reverse process progressively denoises latent motion starting from $\mathbf{z}_m^T \sim N(\mathbf{0}, \mathbf{I})$ via the parameterized Gaussian transition
\begin{equation}
    \begin{aligned} 
    & p_\theta\left(\mathbf{z}_m^{t-1} \mid \mathbf{z}_m^t, y\right)=\mathcal{N}\left(\mathbf{z}_m^{t-1}; \mu_\theta\left(\mathbf{z}_m^t, t, y\right), \sigma_t^2 \mathbf{I}\right), \\
    & \mu_\theta\left(\mathbf{z}_m^t, t, y\right)=\frac{1}{\sqrt{\alpha_t}}\left(\mathbf{z}_m^t-\frac{\beta_t}{\sqrt{1-\bar{\alpha}_t}} \epsilon_\theta\left(\mathbf{z}_m^t, t, y\right)\right),
    \end{aligned}
\end{equation}
where $\epsilon_\theta$ is a trainable autoencoder to approximate the noise. $y$ represents the conditions guiding the generating of motion, which include the pre-obtained content feature $f_{x_0}$ to ensure realism of generated motion, as well as optional class label or text to provide additional control over the motion generation process. We follow \citep{rombach2022high} using a 2D-UNet backbone as $\epsilon_\theta$ with the cross-attention mechanism to incorporate conditions. The training objective is simplified to
\begin{equation}
    \mathcal{L}_{\text {simple }}(\theta)=\left\|\bm{\epsilon} - \epsilon_\theta\left(\mathbf{z}_m^t, t,y\right)\right\|_2^2.
\end{equation}

\begin{figure}[!t]
    \centering
    \includegraphics[width=0.46\textwidth]{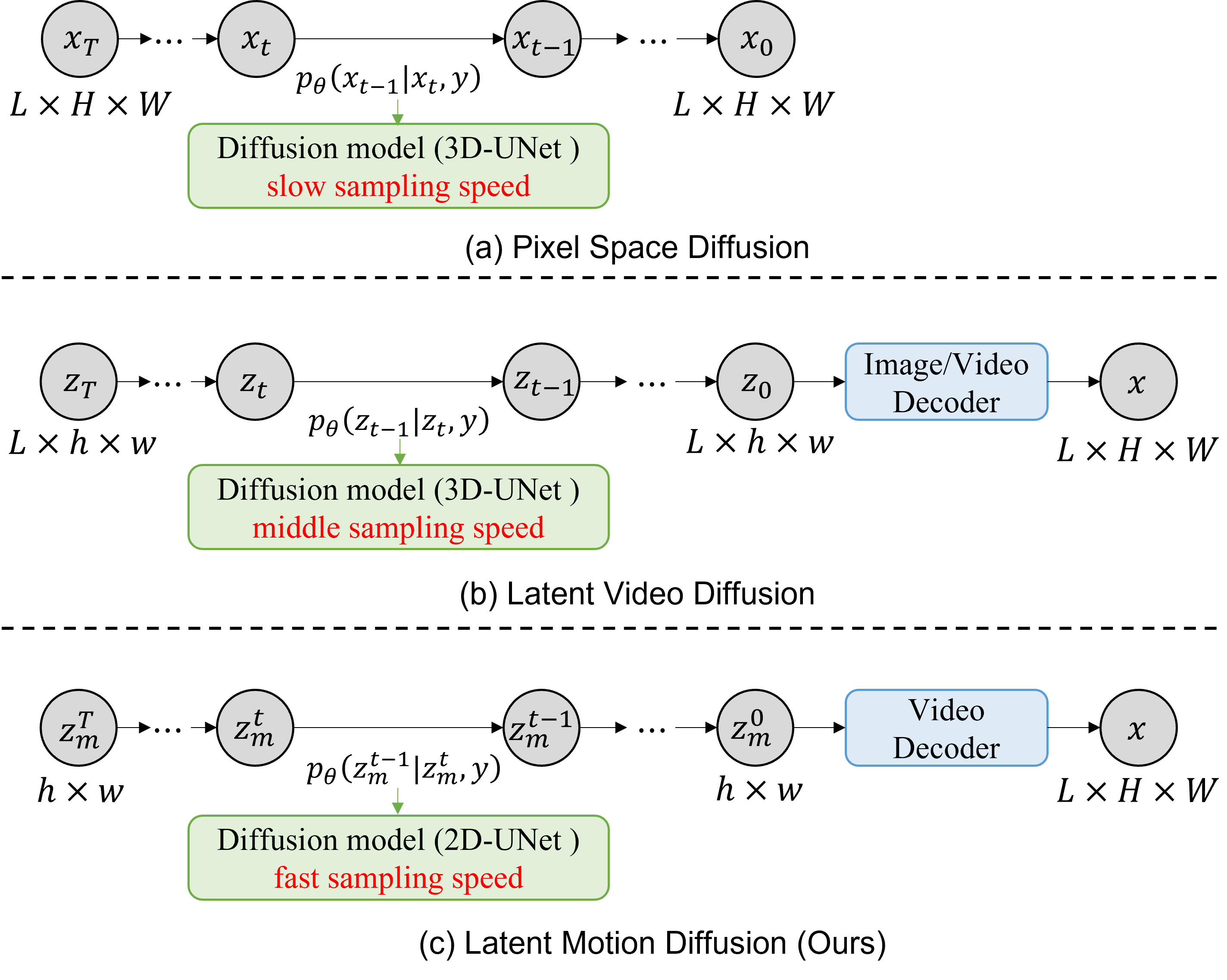}
    \caption{The comparison of sampling process of different video diffusion models. Benefited from low-dimensional diffusion target and 2D-UNet based diffusion model, our latent motion diffusion achieves much faster sampling speed compared to video space diffusion and latent video diffusion. The channel dimension is omitted in all settings for simplicity.}
    \label{fig:paradigms_diffusion}
\end{figure}

During sampling, we adopt the noise schedule from a subsequence with $K$ sampling steps from $[1, 2, \dots, T]$ to improve the sampling speed \citep{nichol2021improved}. Even though, in diffusion models, the generator $\epsilon_\theta$ is recursively executed during sampling. As a result, the computational complexity of $\epsilon_\theta$ can significantly impact the sampling speed, particularly for video data. Therefore, we analyze the sampling efficiency of latent motion diffusion by comparing the computational cost of different video diffusion models as depicted in Fig.\ref{fig:paradigms_diffusion}. Directly generating video pixels with dimension $L \times H \times W$ using pixel space diffusion based on a 3D-UNet model results in the slowest sampling speed. In comparison, by compressing video to latent image/video space with a spatial downsampling ratio $r_s$ and temporal downsampling ratio $r_t$ ($1\leqslant r_{t} < L$), latent video diffusion could reduce the computational complexity by $r_{t} \cdot r_{s}^{2}$ times under the same model architecture and channel size. In contrast, the complexity of our proposed latent motion diffusion, based on a 2D-UNet structure, is at least $L \cdot r_{s}^{2}$ times less than pixel space diffusion which discards the temporal convolution of 3D-UNet. Our latent motion diffusion achieves a computational cost comparable to image diffusion models and the fastest sampling speed among video diffusion paradigms.

\section{Experiments}\label{sec.exp}
We evaluate our method on different image-conditional video generation tasks, including stochastic image-to-video generation, class-guided image-to-video generation and text-image-to-video generation. The experiments are carried out across five benchmark datasets spanning diverse domains such as robotics, landscape, human body/face, and synthetic scenario. These benchmarks encompass a variety of stochastic dynamics and highly controllable motion.

\begin{table*}[t]
\caption{Hyperparameters and training details for MCD-VAE. All models trained on 6 NVIDIA GeForce RTX 3090 GPUs.}
\small
\centering
\begin{tabular}{lccccc}
\toprule
& BAIR & Landscape & NATOPS & MUG & CATER-GEN-v2  \\
\midrule
Resolution & $16 \times 64 \times 64 \times 3$ & $32 \times 128 \times 128 \times 3$ & $32 \times 128 \times 128 \times 3$ & $16 \times 256 \times 256 \times 3$ & $16 \times 128 \times 128 \times 3$ \\
$r_s$ & 4 & 4 & 4 & 4 & 4 \\
$r_t$ & 16 & 32 & 32 & 16 & 16 \\
$z_m$-shape & $16 \times 16 \times 3$ & $32 \times 32 \times 3$ & $32 \times 32 \times 3$ &  $32 \times 32 \times 3$ & $32 \times 32 \times 3$\\
$\mathcal{E}_I$-channels & 128 & 128 & 128 & 128 & 128 \\
$\mathcal{E}_M$-channels & 64 & 64 & 32 & 64 & 64 \\
$\mathcal{D}_V$-channels & 128 & 128 & 128& 128 & 128  \\
Channel Multiplier & 1,2,4 & 1,2,4 & 1,2,4 & 1,2,4 & 1,2,4 \\
$\beta$ & 1e-2 & 1e-2 & 1e-2 & 1e-1 & 1e-2 \\
$N_{params}$ & 165M & 175M & 149M &  175M & 165M \\
Optimizer & Adam & Adam & Adam & Adam & Adam \\
Basic Learning Rate& $\text{5e-5}$ & $\text{5e-5}$ & $\text{5e-5}$ & $\text{5e-5}$ & $\text{5e-5}$\\
Training epochs & 600 & 200 & 500 & 1000 & 200 \\
\bottomrule
\end{tabular}
\label{tab:para_vae}
\end{table*}

\begin{table*}[htbp]
    \caption{Hyperparameters and training details for DMG. All models trained on 6 NVIDIA GeForce RTX 3090 GPUs. The sampling time is measured on a single NVIDIA GeForce RTX 3090 GPU.}
    \small
    \centering
    \begin{tabular}{lccccc}
        \toprule
        & BAIR & Landscape & NATOPS & MUG & CATER-GEN-v2  \\
        \midrule
        $z_m$-shape & $16 \times 16 \times 3$ & $32 \times 32 \times 3$ & $32 \times 32 \times 3$ &  $32 \times 32 \times 3$ & $32 \times 32 \times 3$\\
        Diffusion steps & 1000 & 1000 & 1000 & 1000 & 1000 \\
        Noise Schedule & linear & linear & linear & linear & linear \\
        Base channels & 192 & 192 & 192 & 192 & 192 \\
        Channel Multiplier & 1,2,4 & 1,2,4 & 1,2,4 & 1,2,4 & 1,2,4 \\
        Blocks per resolution & 2 & 2 & 2 & 2 & 2 \\
        Attention Resolution & 1,2,4 & 1,2,4 & 1,2,4 & 1,2,4 & 1,2,4 \\
        Conditioning embedding dimension & 512 & 512 & 512 & 512 & 512 \\
        Conditioning transformer dimension & - & - & 512 & 512 & 512 \\
        $N_{params}$ & 241M & 241M & 282M & 282M & 282M \\
        Dropout & 0.2 & 0.2 & 0.2 & 0.2 & 0.2 \\
        Optimizer & Adam & Adam & Adam & Adam & Adam \\
        Learning Rate& $\text{2e-5}$ & $\text{2e-5}$ & $\text{2e-6}$ & $\text{2e-6}$ & $\text{2e-5}$\\
        Batchsize & 96 & 96 & 96 & 96 & 96 \\
        Training epochs & 400 & 400 & 400 & 600 & 400 \\
        \midrule
        Sampling timesteps & 200 & 200 & 200 & 200 & 200 \\
        Sampling time (per video) & 9.6s & 9.8s & 9.7s & 9.7s & 9.8s \\
        \bottomrule
    \end{tabular}
    \label{tab:para_dmg}
\end{table*}

\subsection{Datasets and Evaluation Metrics}
\noindent\textbf{Datasets:} Here we summarize the datasets used for evaluation, which include BAIR Robot Pushing and Landscape for the stochastic I2V task, NATOPS and MUG for the cI2V task, and CATER-GEN for the TI2V task. 
\begin{itemize}
    \item \textbf{BAIR Robot Pushing} dataset \citep{ebert2017self} consists of videos about a robotic arm randomly pushing a variety of objects in a tabletop setting for real-world interactive agents. It contains about 40k training videos and 256 testing videos. Following the standard protocol \citep{weissenborn2019scaling,rakhimov2021latent,babaeizadeh2021fitvid}, our model generates video with 16 frames given the first image at a resolution of 64 × 64. We generate 100 samples conditioned per image and compare the $100\times256$ videos against 256 testing videos.
    \item \textbf{Landscape} dataset \citep{xiong2018learning} contains about 5k time-lapse videos which have been manually cut into short clips. Those scenes contain a wide range of content, weather, and motion speed, such as cloudy sky with moving clouds, and the starry sky with moving stars. This dataset contains 35,392 clips for training and 2,815 clips for testing. Following previous work \citep{xiong2018learning,zhang2020dtvnet,9577375}, our model generates videos with 32 frames at a resolution of 128 × 128 for fair comparison. 
    \item \textbf{NATOPS} aircraft handling signal dataset \citep{5771448} comprises 24 gestures performed by 20 subjects. Each gesture is repeated 20 times by one subject, resulting in a total of 9,600 videos. We use the same training and testing split as \citep{Ni_2023_CVPR} that involves 4,800 clips from 10 subjects for training and 4,800 clips from the remaining 10 subjects for testing. The model is trained on videos with 32 frames at a resolution of 128 × 128. During evaluation, we generate 10,000 videos and compare against testing test following \citep{Ni_2023_CVPR}.
    \item \textbf{MUG} facial expression dataset \citep{5617662} contains 1,009 videos of 52 subjects performing 7 different expressions. We use the same split as \cite{Ni_2023_CVPR} with 465 video from 26 subjects for training and 544 videos from the remaining 26 subjects for testing. To align the video setting with compared methods, we adopt two video resolutions: 32 frames at a resolution of 128 × 128 to compare with \cite{Ni_2023_CVPR}, and 16 frames at a resolution of 256 × 256 following \cite{Ni_2024_CVPR}.
    \item \textbf{CATER-GEN-v2} \citep{hu2022make} is synthetic dataset built in a 3D environment with realistic lighting and shadows. Each video contains 3$\sim$8 objects and 2 movements randomly selected from four atomic actions. Each object has four attributes that are randomly chosen from the respective set of shape, size, color, and material. The text descriptions are highly motion related and introduce the uncertainty of action and referring expression. There are 24k training video-text pairs and 6k testing pairs, and we generate videos at a resolution of 128x128 and with a sequence length of 16 frames.
\end{itemize}

\noindent\textbf{Implementation Details:} The MCD-VAE downsamples the spatial dimension of latent motion by a factor of $r_{s}=4$. The temporal downsampling ratio $r_t$ is set to be equal with the frame length of input video with $r_{t}=\left\{16, 32\right\}$, which further aligns with the experimental setting in compared methods. The channel size $d$ of latent motion $z_m$ is set to be 3 which derived from the corresponding abaltion study in Sec.\ref{sec:ablation_d}.
The DMG applies similar 2D-UNet backbone with cross-attention based conditional blocks as \citep{rombach2022high}. 
The MCD-VAE and DMG are both trained on the same training set of corresponding dataset without the use of external data, although the expressiveness and reconstruction quality of MCD-VAE could be further improved benefiting from extensive video data. During sampling, the latent motion is generated with a subsequence of $K=200$ timesteps. More training details are listed in Tab.\ref{tab:para_vae} and Tab.\ref{tab:para_dmg}.

\noindent\textbf{Evaluation Metrics:} To evaluate the visual quality of generated videos, we first follow the metrics used in previous work \citep{rakhimov2021latent, yan2021videogpt, 9577375} for stochastic image-to-video generation, including the image-level perceptual similarity metrics Fr\'echet Inception Distance (FID) \citep{heusel2017gans} and Learned Perceptual Image Patch Similarity (LPIPS) \citep{dosovitskiy2016generating}, as well as the video-level metric Fr\'echet-Video-Distance (FVD) \citep{unterthiner2018towards}.

Considering the limitation of these metrics on evaluating motion quality, we have carefully investigated applicable evaluation metrics with the consideration of different tasks and datasets. For stochastic image-to-video generation, we additionally compare the motion smoothness, dynamic degree, temporal flickering, and background consistency from VBench \citep{Huang_2024_CVPR}. The motion smoothness measures the smoothness of generated motions via the motion priors in the video frame interpolation model, while dynamic degree estimates the degree of dynamics by measuring optical flow strengths. To evaluate the temporal consistency, temporal flickering computes the mean absolute different across frames, and background consistency calculates CLIP feature similarity across frames. The higher score indicates relatively better performance for the corresponding dimension. 

For the class-guided image-to-video generation, we add two variants of FVD flollowing \cite{Ni_2023_CVPR, Ni_2024_CVPR}, the class conditional FVD (cFVD) and subject conditional FVD (sFVD), to evaluate the video-text alignment and subject consistency, respectively. cFVD and sFVD compare the distance of feature distributions between real and generated videos under the same class condition or the same subject image condition. Both the mean and standard variance of cFVD and sFVD are reported to compare the distribution over different classes and subjects.

For the text-image-to-video generation, we add the action precision and referring expression precision metrics, specially designed for CATER-GENs as mentioned in \cite{10148799}, to evaluate the motion accuracy. The action precision explicitly measures the movements accuracy through the detected object coordinates, while referring expression precision assesses the correctness of the moving entities by comparing their attributes in the generated videos to those described in the text. The range of these two metrics fall into $[0,1]$ which the larger value indicates better performance.

\begin{figure}[t]
    \centering
    \includegraphics[width=0.48\textwidth]{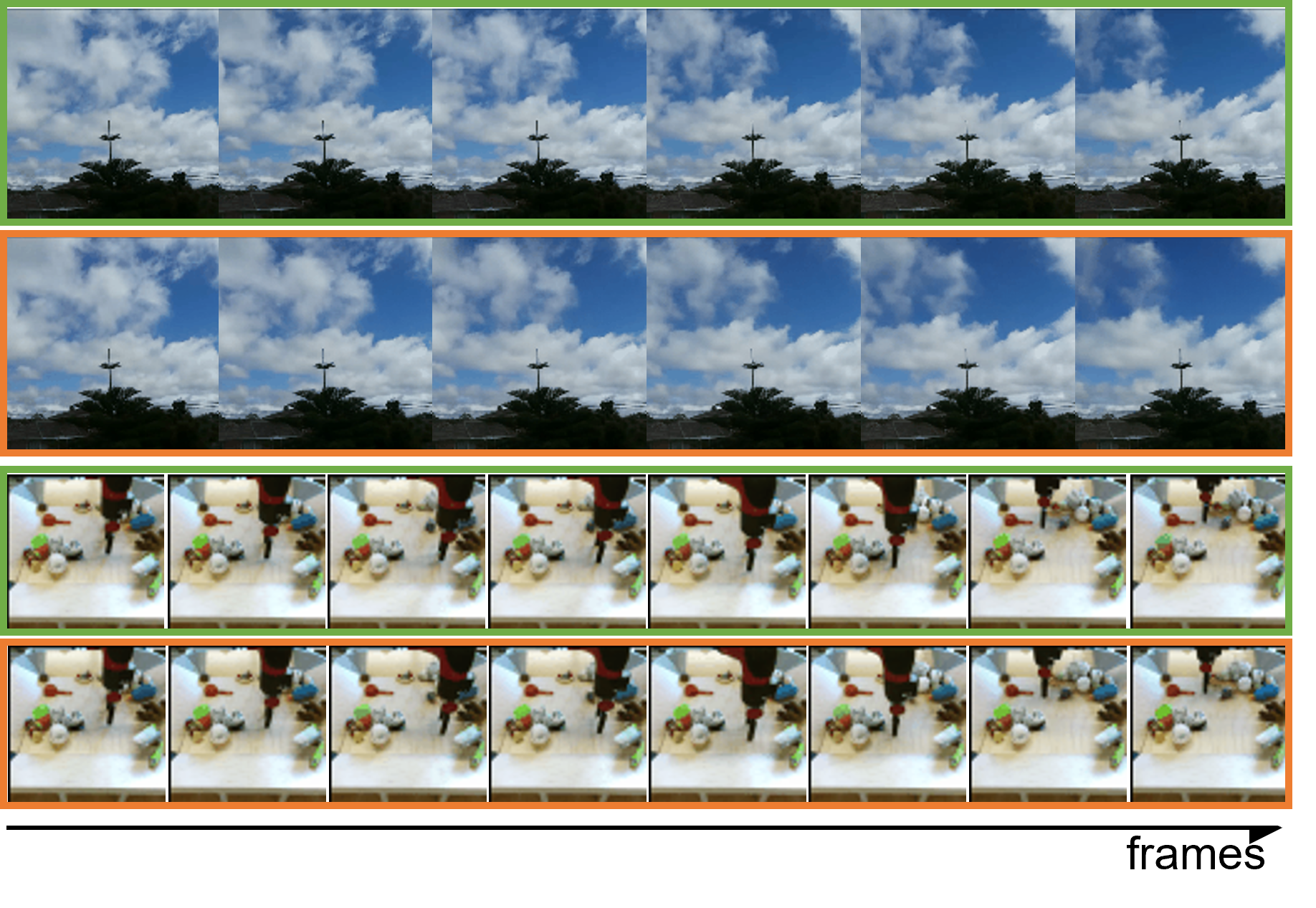}
    \caption{The reconstruction results of MCD-VAE on Landscape and BAIR datasets. The original videos are contained in green boxes, while reconstructed videos in orange boxes.}
    \label{fig:mcd-vae}
\end{figure}

\begin{table}[t]
    \caption{The quantitative results of reconstruction performance of MCD-VAE on different datasets.}
    \small 
    \centering
    \label{tab:mcd-vae}
    \begin{tabular}{c|cccccc}
        \toprule
        Datasets & $r_{s}$ & $r_{t}$ & $d$ & LPIPS$\downarrow$ & FID$\downarrow$ & FVD$\downarrow$ \\
        \midrule
        BAIR & 4 & 16 & 3 & 0.02 & 5.52 & 36.05 \\
        Landscape & 4 & 32 & 3 & 0.09 & 2.48 & 58.59 \\
        NATOPS & 4 & 32 & 3 & 0.04 & 19.41 & 78.32 \\
        MUG & 4 & 16 & 3 & 0.05 & 5.49 & 19.26 \\
        CATER-GEN-v2 & 4 & 16 & 3 & 0.02 & 1.39 & 2.18 \\
        \bottomrule
    \end{tabular}%
\end{table}

\begin{table}[t]
    \centering
    \small 
    \caption{The comparison of reconstruction performance on BAIR dataset.}
    \begin{tabular}{c|ccccc}
        \toprule
        Methods & SSIM $\uparrow$ & PSNR $\uparrow$& FID $\downarrow$ & LPIPS $\downarrow$& FVD $\downarrow$\\
        \midrule
        VQGAN & 0.97 & 31.63 & 8.29 & 0.02 & 42.80 \\
        MCD-VAE & 0.97 & 31.64 & 5.52 & 0.02 & 36.05 \\
        \bottomrule
    \end{tabular}
    \label{tab:mcd-vae-vqgan}
\end{table}

\subsection{Reconstruction Ability of MCD-VAE}\label{sec:exp_vae}
MCD-VAE is designed to perform motion decomposition and compression, as well as motion-content fusion and pixel-level reconstruction. Therefore, here we evaluate the decomposition and reconstruction capability of MCD-VAE through quantitative and qualitative analysis.

\subsubsection{Reconstruction Capability }
To evaluate the quality of reconstruction, we compare real videos with reconstructed videos which are synthesized from appearance feature of the first frame and latent motion of the video in Fig.\ref{fig:mcd-vae}. The results show that the latent motion successfully captures fine-grained movements such as the cloud deformation under a large compression ratio ($r_t=32, r_s=4$). We also list the quantitative results on the testing set of different datasets in Tab.\ref{tab:mcd-vae} to provide a reference for the subsequent video generation performance, as the reconstruction performance acts as an upper bound for the generation results. Additionally, to assess the adequacy of the reconstruction performance for satisfactory video generation, we compare it with VQGAN \citep{esser2021taming}, a common image autoencoder used in both image and video generation approaches, in Tab.\ref{tab:mcd-vae-vqgan}. VQGAN employs a distinct compression strategies for spatial and temporal dimensions by individually encoding each frame and compressing solely spatial information. However, the results indicate that MCD-VAE achieves comparable performance in terms of frame-level metrics, while also demonstrating superior performance in video-level metric. This finding further reinforces the notion that motion can be decomposed and highly compressed while still retaining sufficient expressiveness to recover the video.

\begin{figure*}[!th]
    \centering
    \includegraphics[width=1.\textwidth]{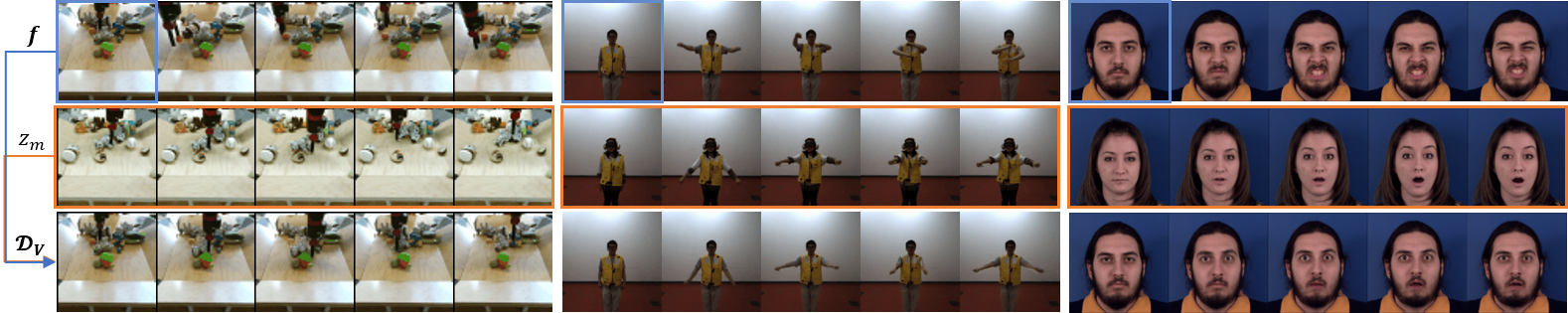}
    \caption{Visualization of latent motion transfer to evaluate the efficacy of decomposition. The synthesized video (the third row) is reconstructed using the content feature $\mathbf{f}$ extracted from the first frame of one video (the first row), combined with the latent motion feature $z_m$ extracted from the other video (the second row).}
    \label{fig:decompose}
\end{figure*}

\begin{figure*}[htbp]
    \centering
    \includegraphics[width=0.95\textwidth]{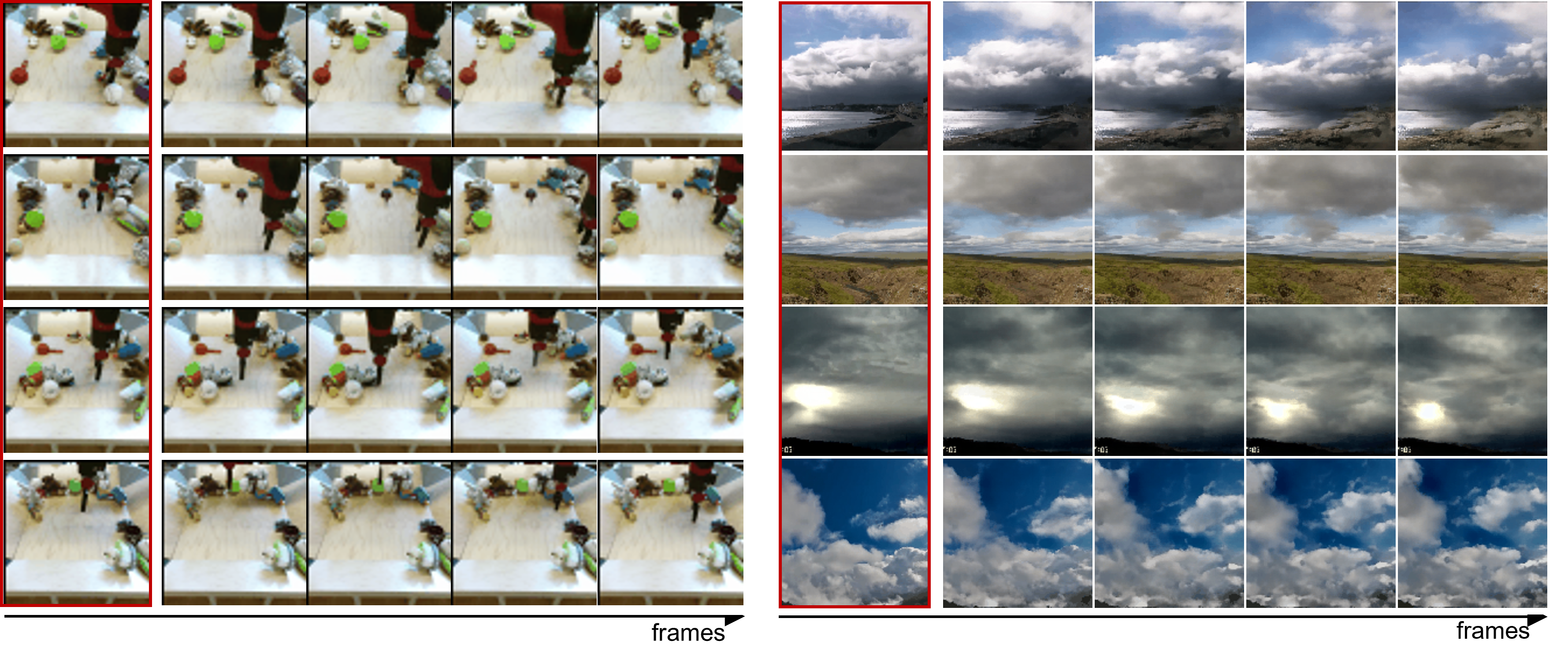}
    \caption{The generated samples on BAIR dataset and Landscape dataset. The given frame $x_0$ is highlighted with red box. We show the $4_{th}, 8_{th}, 12_{th}, 16_{th}$ frames for BAIR dataset and $8_{th}, 16_{th}, 24_{th}, 32_{th}$ frames for Landscape dataset, respectively.}
    \label{fig:generate_stochastic}
\end{figure*}

\subsubsection{Decomposition Capability }
Measuring the efficacy of decomposition quantitatively presents a challenge, yet we can evaluate whether the motion is cleanly decoupled from content by combining the latent motion derived from video $x^1$ with content features extracted from a different video $x^2$. If the motion and content are successfully decoupled, the synthesized video will exhibit the motion patterns of $x^1$, but preserve the visual appearance of $x^2$. Otherwise, the synthesized video may leak appearance from $x^1$ or display movement inconsistent with $x^1$. To demonstrate that our model has successfully decoupled appearance and motion, we present visualizations of latent motion transfer on various datasets in Fig.\ref{fig:decompose}. The synthesized videos depicted in the third row is reconstructed using the content feature $\mathbf{f}$ extracted from the first frame of video in the first row, combined with the latent motion feature $z_m$ extracted from the video in the second row. Fig.\ref{fig:decompose} shows that the movements in the synthesized video are consistent with video in the second row, while the visual content matches the video in the first row. These outcomes confirm that MCD-VAE has successfully decomposed motion and appearance information.

\begin{table*}[htbp]
    \caption{Quantitative evaluation on motion quality and temporal consistency on the Landscape dataset.}
    \small
    \centering
    \begin{tabular}{c|cccc}
        \toprule
        Method & Motion Smoothness & Dynamic Degree & Temporal Flickering & Background Consistency \\
        \midrule
        cINN \citep{9577375} & 99.41\% & 40.49\% & 99.14\% & \textbf{97.46}\% \\
        DynamiCrafter \citep{xing2023dynamicrafter} & 96.89\% & \textbf{94.60}\% & 95.30\% & 93.96\% \\
        \midrule
        LaMD (Ours) & \textbf{99.47}\% & 55.13\% & \textbf{99.15}\% & 96.72\% \\
        \bottomrule
    \end{tabular}
    \label{tab:landscape_motion}
\end{table*}

\subsection{Video Generation Results}
\subsubsection{Stochastic image-to-video generation}
To evaluate the performance of LaMD on the stochastic I2V generation task, where videos with stochastic dynamic movements are generated from a single image, we first showcase the generated samples on the BAIR and Landscape datasets in Fig.\ref{fig:generate_stochastic}. These samples exhibit indistinguishable appearance quality from the given image, aided by the incorporation of multi-scale content features that capture rich details. And the generated videos successfully depict reasonable cloud deformation and maintain consistency throughout long-range movements. Additionally, we show the video diversity in Fig.\ref{fig:diverse} by generating multiple videos from the same image, which exhibit realistic and diverse motion.

To further assess our model's performance, we compare the quantitative results with the state-of-the-art work in Tab.\ref{tab:bair_sota} and Tab.\ref{tab:landscape_sota}. The results highlight that our model outperforms state-of-the-art methods significantly in terms of overall video quality, as indicated by the FVD metric. 
For Landscape benchmark that comes from the realistic natural scenes, we additionally compare our model with two baseline methods on motion smoothness, dynamic degree, temporal flickering, and background consistency in Tab.\ref{tab:landscape_motion}. Overall, our method achieves superior performance on motion smoothness and temporal flickering than all baseline models. Compared to the same close-domain I2V generation utilizing limited training data as cINN \citep{9577375}, our method can generate videos with larger motions, with only a marginal reduction in background consistency. In contrast, when comparing against the open-domain I2V method, we found that DynamiCrafter \citep{xing2023dynamicrafter} demonstrates higher degree of dynamics, attributable to the diverse motion patterns present in its extensive training dataset, including various camera movements. Consequently, our method outperforms the open-domain I2V approach in terms of motion smoothness and temporal consistency.

\begin{table}[t]
    \caption{Quantitative evaluation compared to the state-of-the-art on the BAIR dataset.}
    \centering
    \small
    \setlength{\tabcolsep}{5mm}{
        \begin{tabular}{c|c}
            \toprule
            Method & FVD $\downarrow$ \\
            \midrule
            LVT \citep{rakhimov2021latent} & 125.8 \\
            DVD-GAN \citep{clark2019adversarial} & 109.8 \\
            VideoGPT \citep{yan2021videogpt} & 103.3 \\
            cINN \citep{9577375} & 99.6 \\
            Video Transformer \citep{weissenborn2019scaling} & 94.0 \\
            FitVid \citep{babaeizadeh2021fitvid} &  93.6 \\
            N{\"U}WA \citep{wu2022nuwa} & 86.9 \\
            MMVG \citep{Fu_2023_CVPR} & 85.2 \\
            MOSO \citep{sun2023moso} & 83.6 \\
            VDM \citep{ho2022video} & 66.9 \\
            \midrule
            LaMD (Ours) & \textbf{57.0} \\
            \bottomrule
    \end{tabular}}
    \label{tab:bair_sota}
\end{table}

\begin{table}[t]
    \caption{Quantitative evaluation compared to the state-of-the-art methods on the Landscape dataset. The results of methods for comparison are quoted from corresponding paper or reported by \citep{9577375}.}
    \small
    \centering
    \setlength{\tabcolsep}{2.8mm}{
        \begin{tabular}{c|ccc}
            \toprule
            Method & LPIPS $\downarrow$ & FID $\downarrow$ & FVD $\downarrow$ \\
            \midrule
            MDGAN \citep{xiong2018learning} & 0.49 & 68.9 & 385.1 \\
            DTVNet \citep{zhang2020dtvnet} & 0.35 & 74.5 & 693.4 \\
            DL \citep{logacheva2020deeplandscape} & 0.41 & 41.1 & 351.5 \\
            AL \citep{endo2019animating} & 0.26 & 16.4 & 307.0 \\
            cINN \citep{9577375} & \textbf{0.23} & 10.5 & 134.4 \\
            \midrule
            LaMD (Ours) & 0.26 & \textbf{8.67} & \textbf{100.7} \\
            \bottomrule
    \end{tabular}}
    \label{tab:landscape_sota}
\end{table}

\subsubsection{Class-guided image-to-video generation} 
\begin{figure*}[htbp]
    \centering
    \includegraphics[width=1.0\textwidth]{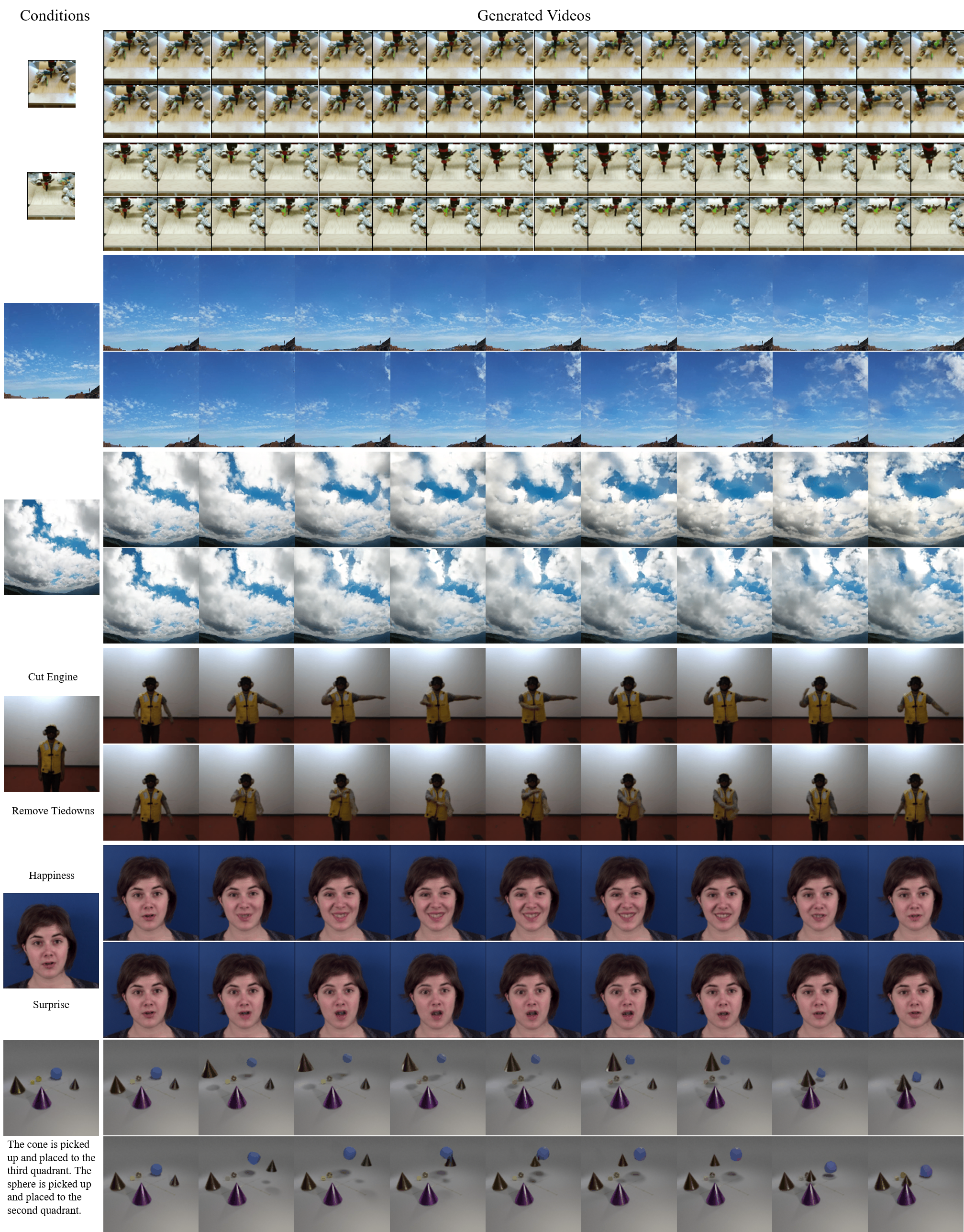}
    \caption{Diverse video generation samples (right) under given conditions (left). For the NATOPS and MUG datasets, we show two distinct generated examples for each, using the same given image but varying the action classes to demonstrate the model's controllability. Fow other datasets, we exhibit two generated samples under same condition to illustrate the diversity of generated motion.}
    \label{fig:diverse}
\end{figure*}

Apart from generating stochastic movements, we also evaluate the ability of generating coarse-grained motion under the guidance of an action label on the NATOPS and MUG dataset.
For quantitative results, we adopt FVD, sFVD, and cFVD to evaluate respective overall video quality, video-text alignment, and subject consistency for class-guided I2V generation.
The results on these two datasets are reported in Tab.\ref{tab:natops_sota} and Tab.\ref{tab:mug_sota}, respectively. The quantitative results demonstrate that our method outperforms baseline models that generate videos in latent image/video space and achieves competitive results with LFDM \citep{Ni_2023_CVPR}, which generates optical flows in latent space. Moreover, our proposed method exhibits significantly lower standard variance in sFVD and cFVD across different subjects or classes, along with faster sampling speeds, as discussed in Section \ref{sec:speed}. Additionally, DynamicCrafter \citep{xing2023dynamicrafter} and TI2V-Zero \citep{Ni_2024_CVPR} are the recent open-domain TI2V generation methods that are built upon pre-trained large-scale T2V models. Our model demonstrates a marked enhancement in both video quality and consistency of conditions, coupled with a much faster sampling rate. The generated samples are shown in Fig.\ref{fig:diverse}. The subjects in generated videos well maintain the appearance in the first frame, while perform the consistent expression with the action condition.

\begin{figure*}[t]
    \centering
    \includegraphics[width=1.\textwidth]{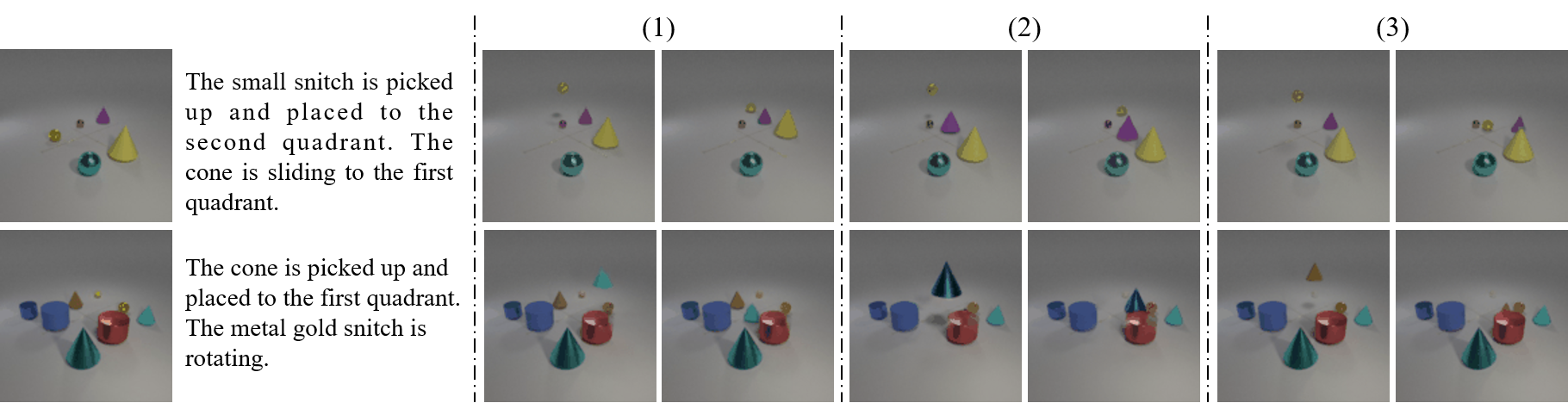}
    \caption{The generated samples on CATER-GEN-v2 dataset under diverse video generation. Given an image and description in the left column, we generate three videos and display the 8th and 16th frames of each video in the right three columns.}
    \label{fig:cater}
\end{figure*}

\begin{table*}[t]
    \caption{Quantitative evaluation compared to the state-of-the-art on the MUG dataset. The subscript 128 and 256 represents the spatial resolution $128\times128$ and $256\times256$ of generated videos, respectively.}
    \centering
    \small
    \setlength{\tabcolsep}{3mm}{
        \begin{tabular}{c|ccc}
            \toprule
            Method & FVD $\downarrow$ & cFVD $\downarrow$ & sFVD $\downarrow$ \\
            \midrule
            LDM$_{128}$ \citep{rombach2022high} & 126.28 & 208.03$\pm$64.86 & 241.49$\pm$75.18 \\
            LFDM$_{128}$ \citep{Ni_2023_CVPR} & 32.09 & 84.52$\pm$24.81 & 114.33$\pm$42.62 \\
            LaMD$_{128}$ (Ours) & 38.62 & 88.81$\pm$9.80  & 126.45$\pm$40.31 \\
            \midrule
            DynamiCrafter$_{256}$ \citep{xing2023dynamicrafter} & 1094.72  & 1223.89$\pm$105.94 & 1359.86$\pm$257.73 \\
            TI2V-Zero$_{256}$ \citep{Ni_2024_CVPR} & 180.09 & 252.77$\pm$39.02 & 267.17$\pm$74.72 \\
            LaMD$_{256}$ (Ours) & \textbf{49.62} & \textbf{103.97$\pm$ 23.73} & \textbf{129.24$\pm$ 36.47} \\
            \bottomrule
    \end{tabular}}
    \label{tab:mug_sota}
\end{table*}

\begin{table}[htbp]
    \caption{Quantitative evaluation compared to the state-of-the-art on the NATOPS dataset.}
    \centering
    \small
    \setlength{\tabcolsep}{1mm}{
        \begin{tabular}{c|ccc}
            \toprule
            Method & FVD $\downarrow$ & cFVD $\downarrow$ & sFVD $\downarrow$ \\
            \midrule
            LDM \citep{rombach2022high} & 344.81 & 627.84$\pm$169.52 & 623.13$\pm$320.85\\
            LFDM \citep{Ni_2023_CVPR} & \textbf{195.17} & \textbf{423.42}$\pm$117.06 & \textbf{369.93}$\pm$\textbf{159.26} \\
            \midrule
            LaMD (Ours) & 196.67 & 432.62$\pm$103.27  & 386.12$\pm$160.80 \\
            \bottomrule
    \end{tabular}}
    \label{tab:natops_sota}
\end{table}

\subsubsection{Text-image-to-video generation}
TI2V is a recently proposed task that involves generating videos from a given image and fine-grained text descriptions. In contrast to I2V and cI2V, TI2V usually requires greater control over the motion, which is specified in detail through text descriptions. To enable this, we concatenate the text embeddings extracted by a text encoder with the content features of the given image as multi-modal conditions in DMG. Hence, the latent motion could retrieve information from both image appearance and text descriptions.

For quantitative evaluation, since the text descriptions in CATER-GEN-v2 are closely linked to fine-grained motion, making it challenging for most current metrics to accurately assess the precision of detailed movements. Especially, the descriptions encompass ambiguous referring expressions and motions, allowing for the existence of multiple videos that satisfy the given text. Therefore, we utilize the action precision and referring expression precision metrics, specially designed for CATER-GEN as mentioned in \cite{10148799}, to evaluate the motion accuracy.
According to the quantitative results presented in Tab.\ref{tab:cater_sota}, LaMD significantly ourperforms existing methods not only on the visual quality metrics LPIPS, FID, and FVD, but also on the motion accuracy with $P_{act}$ and $P_{re}$. This signifies the effectiveness of LaMD in generating highly controllable and diverse motion, benefited from the framework of latent motion generation.
Fig.\ref{fig:cater} showcases the generated videos on CATER-GEN-v2. With the given image and text as inputs, we generate multiple videos which show satisfying diversity while maintaining coherent movements as specified in the text.

\begin{table}[t]
    \caption{Quantitative results on the CATER-GEN-v2 under diverse video generation.}
    \centering
    \small
    \setlength{\tabcolsep}{1.1mm}{
    \begin{tabular}{c|ccc|cc}
        \toprule
        Model & LPIPS$\downarrow$ & FID$\downarrow$ & FVD$\downarrow$ & $P_{act}$ $\uparrow$ & $P_{re}$ $\uparrow$\\
        \midrule
            MAGE \citep{hu2022make} & 0.26 & 39.38 & 69.44 & 0.629 & 0.602 \\
        MAGE+ \citep{10148799} & \textbf{0.09} & 3.44 & 23.43 & 0.735 & 0.707 \\
        LaMD (Ours) & \textbf{0.09} & \textbf{2.26} & \textbf{5.77} & \textbf{0.840} & \textbf{0.748} \\
            \bottomrule
    \end{tabular}}
    \label{tab:cater_sota}
\end{table}

\begin{table*}[htbp]
    \centering
    \small
    \caption{The comparison of sampling time per video. All diffusion models use 200 sampling steps and are tested on a single NVIDIA GeForce RTX 3090 GPU.}
    \setlength{\tabcolsep}{4mm}{
    \begin{tabular}{@{}cccccc@{}}
    \toprule
    Diffusion Space   & Methods  & Default $N_{params}$    & $N_{frames}$ & Resolution & Sampling Time (per video) $\downarrow$ \\ \midrule
    Latent image space & LDM \citep{rombach2022high} & $\sim$600M & 1  & 128x128 & 9.8s \\ \midrule
    \multirow{2}{*}{Pixel space}  & \multirow{2}{*}{VDM \citep{ho2022video}}  & \multirow{2}{*}{$\sim$296M} & 16  & 32$\times$32 & 263s   \\
                                        &     &    & 16     & 128$\times$128    & 80min    \\ \midrule
    \multirow{2}{*}{Latent video space}  & \multirow{2}{*}{LVDM \citep{he2022latent}} & \multirow{2}{*}{$\sim$500M} & 16 & 128$\times$128& 13.8s \\
                                        &   &  & 128  & 128$\times$128    & 26.6s    \\ \midrule
    \multirow{5}{*}{Latent motion space} & \multirow{2}{*}{LFDM \citep{Ni_2023_CVPR}} & \multirow{2}{*}{$\sim$100M} & 16 & 128$\times$128  & 10.2s  \\
                                        &   &    & 128 & 128$\times$128& 40.1s  \\ \cmidrule(l){2-6} 
                                        & \multirow{3}{*}{LaMD (Ours)} & \multirow{3}{*}{$\sim$400M} & 16 & 128$\times$128 & 9.7s\\
                                        &   &   & 128 & 128$\times$128 & 10.7s    \\
                                        &   &   & 128 & 256$\times$256 & 12.8s \\ \bottomrule
    \end{tabular}}
    \label{tab:time}
\end{table*}

\subsection{Model Analysis}\label{sec:exp_analysis}
\subsubsection{Sampling Time}\label{sec:speed}
To analyse the sampling time of different paradigms, we select four recent diffusion-based methods for comparison, including LDM \citep{rombach2022high}, VDM \citep{ho2022video}, LVDM \citep{he2022latent}, and LFDM \citep{Ni_2023_CVPR}. LDM is a popular image generation method that uses a 2D-UNet-based diffusion model on latent image space, assisted by an image autoencoder. On the other hand, VDM, LVDM, LFDM empoly 3D-UNet-based diffusion models on pixel space, latent video space and latent motion space, respectively. The sampling time results are reported in Tab.\ref{tab:time}. For fair comparison, we set the sampling step as 200 and test all models on the same single NVIDIA GeForce RTX 3090 GPU. Given that sampling time is closely correlated with model size when employing the similar UNet architecture, we also provide the default model size for reference.
The results indicate that generating directly in pixel space, as in the case of VDM, requires a considerable amount of sampling time. In contrast, LVDM utilizes a video autoencoder and applies the diffusion model in latent video space, followed by an interpolation 3D-UNet for longer video generation. This approach significantly reduces sampling time. However, even with a separate temporal interpolation model, its sampling time for a long video with 128 frames still remains much higher than ours (26.6s v.s. 10.7s), while our method allows for an improved channel size $d$ to boost motion capacity without an increase in sampling time. Moreover, the recent LFDM, similar to ours method, first extracts motion representation from video and then generates it in latent motion space. However, LFDM represents motion as a flow map between two frames, necessitating a 3D-UNet diffusion model to generate a flow sequence for video warping. Although the LFDM has a similar time cost for a 16-frames video with a much smaller model compared to ours, the sampling time significantly increases as the video length increases. 
Lastly, it is worth mentioning that sampling a video with our LaMD achieves a comparable speed to sampling an image with LDM. Moreover, even when the frame length and resolution are increased, the sampling time only exhibits a minor increase. This exemplifies the considerable potential of LaMD in minimizing computational expenses in practical applications.

\begin{table}[t]
    \centering
    \small
    \caption{Ablation study of motion capacity in MCD-VAE on the BAIR dataset.}
    \setlength{\tabcolsep}{2.5mm}{
    \begin{tabular}{c|c|ccccc}
        \toprule
        $d_t$ & $d$ & PSNR$\uparrow$ & SSIM$\uparrow$ & FID$\downarrow$ & LPIPS$\downarrow$ & FVD$\downarrow$ \\
        \midrule
        1 & 1 & 29.88 & 0.958 & 6.24 & 0.03 & 41.91 \\
        1 & 2 & 29.73 & 0.955 & 6.26 & 0.03 & 42.96 \\
        1 & 3 & \textbf{31.64} & \textbf{0.969} & \textbf{5.52} & \textbf{0.02} & \textbf{36.05} \\
        1 & 4 & 31.32 & 0.967 & 5.81 & \textbf{0.02} & 38.84 \\
        \midrule
        2 & 2 & 31.63 & 0.969 & 5.55 & 0.02 & 37.93 \\
    \bottomrule
    \end{tabular}}
    \label{tab:mcd-vae-d}
\end{table}

\begin{table*}[t]
    \centering
    \small
    \caption{The trade-off between latent-space size \& reconstruction \& generation.}
    \setlength{\tabcolsep}{6mm}{
    \begin{tabular}{ccc|ccc|ccc}
        \toprule
        \multicolumn{3}{c|}{Latent-space size} & \multicolumn{3}{c|}{Reconstruction} & \multicolumn{3}{c}{Generation} \\
        Size & $\beta$ & d & FID & LPIPS & FVD & FID & LPIPS & FVD \\
        \midrule
        larger & e-5 & 4 & 2.11 & 0.08 & 37.35 & 10.3 & 0.29 & 127.1\\
        smaller & e-2 & 3 & 2.48 & 0.09 & 58.59 & 8.6 & 0.26 & 100.7 \\
        \bottomrule
    \end{tabular}}
    \label{tab:tradeoff}
\end{table*}

\begin{figure*}[htbp]
    \centering
    \includegraphics[width=1.\textwidth]{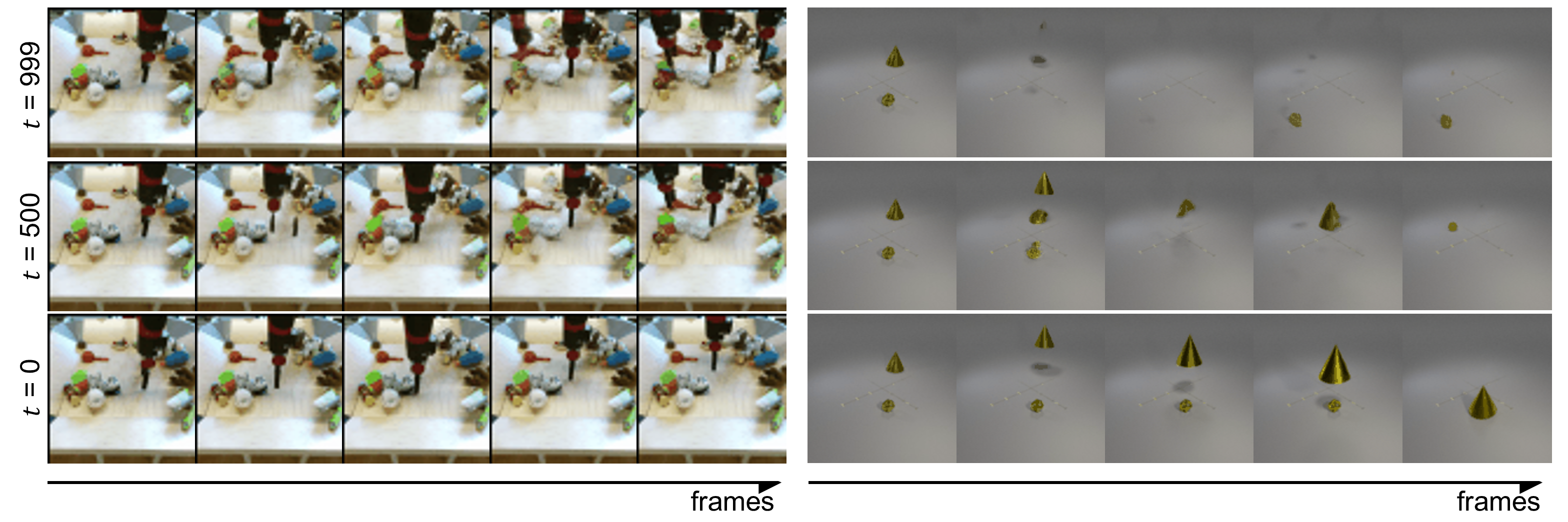}
    \caption{The generated videos at different timesteps during sampling, where $t=999$ corresponds to synthesized videos based on random latent motion and $t=0$ represents final generated videos. (The text condition provided in the example shown in the right column is ``The cone is picked up and containing the snitch''.)}
    \label{fig:diffusion}
\end{figure*}

\subsubsection{Ablation Study of Motion Capacity}\label{sec:ablation_d}
Our efficient generation hinges on compressing motion into a compact latent representation, gaining its advantage from the removal of temporal dimension. 
The fully compression on temporal dimension in MCD-VAE should allow for the transfer of motion information from the temporal into the channel dimension in the latent space without compromising performance. To demonstrate this, we develop a variant of MCD-VAE with fewer temporal up/down-sampling layers, thereby preserving the temporal dimension. The last two rows in Tab.\ref{tab:mcd-vae-d} compare the reconstruction performance with and without the temporal dimension $d_t$ of latent motion under the same capacity. The results show that discarding the temporal dimension does not detrimentally affect performance. More importantly, a critical advantage of removing temporal dimension of latent motion is the facilitation of a 2D-based diffusion model over a 3D-UNet one, which effectively reduces the computational complexity and improves sampling speed. This design makes the model’s inference burden comparable to that of image diffusion models, which has been demonstrated in Sec.\ref{sec:speed}.

In order to explore the necessary capacity for latent motion, we then conduct an ablation study on the channel size $d$ and compare the reconstruction performance in Tab.\ref{tab:mcd-vae-d} (noted that the $\beta$ to control the bottleneck size remains unchanged). Even with a small channel size (e.g. $d=1$), MCD-VAE is able to achieve satisfactory reconstruction performance with high PSNR and SSIM scores, suggesting that there is significant redundancy in the motion information presented in videos. As we increase the channel size, the reconstruction performance improves benefiting from the richer motion information captured, and then slightly attenuates due to the redundancy and potential overfitting. Ultimately, we settle on a channel size of $d=3$, which is consistent with the channel size used in image diffusion model like LDM \citep{rombach2022high}, allowing us to decrease the computational complexity similar to image generation. This finding highlights the potential of motion decomposition and compression in the fields of video compression and generation.

\subsubsection{Trade-off analysis} 
The size of the latent space of motion representation has a dual impact on both the reconstruction performance of MCD-VAE and generation performance of LaMD. In order to investigate the potential trade-off between the latent-space size, reconstruction, and generation, we conduct an experiment on the Landscape dataset in Tab.\ref{tab:tradeoff}. Similar to the findings on the BAIR dataset, using a smaller latent space by reducing the channel size and strengthening KL constraint leads to inferior reconstruction performance of MCD-VAE. However, a more compact latent space has the advantage of alleviating the burden of training a generative model, thereby improving the generation performance of LaMD. 
The similar trade-off also exists in other two-stage paradigms. It provides valuable insights for selecting the appropriate hyper-parameters.

\subsubsection{Visualization of sampling process}
We visualize the videos based on generated motion at different timesteps $t$ during sampling process in Fig.\ref{fig:diffusion} to explore the essence of latent motion. In the first row, videos are generated based on randomly sampled motion representation from normal Gaussian distribution, which is the initial state of latent motion when $t=999$. The corresponding synthesized videos only involve chaotic movements of subjects (robot arm in BAIR and objects in CATER-GENs) while the appearance of background is well maintained, which suggests that MCD-VAE can effectively extract motion patterns from the data distribution and also validates the decomposition ability of motion. 
During recursive sampling process, the movements in generated videos are more orderly in pace with the gradual denoising of latent motion, and finally become natural based on final state of latent motion at timestep $t=0$ in the last row. It validates that generated motion is progressively refined torwards the direction of fitting content condition.

\section{Conclusion and Discussion}\label{sec:discuss}
In this section, we summarize the contributions and findings of this paper and discuss the limitations for future researches. 
In summary, we introduced a novel paradigm and corresponding framework LaMD for video generation. By decomposing and compressing motion from videos, the video generation is refactored into motion generation and video reconstruction. Specifically, MCD-VAE is designed to highly compress motion representation and reconstruct videos through motion-content fusion. The DMG is applied to progressively generate the decomposed motion on continuous latent space. Our framework supports image-conditional video generation tasks and achieves much faster sampling speed than other video diffusion models. The experiments demonstrate that LaMD is capable of generating high-quality videos with a wide range of motions, from stochastic dynamics to highly controllable movements. 

While our approach has shown promising performance across multiple datasets and tasks, it is important to acknowledge its limitations. 
First, the decomposition of latent motion is primarily effective for short videos where the content does not undergo significant changes. According to design of our framework, the latent motion encompasses information not present in the first frame, which may entangle partial appearence features from emerging elements for long videos accompanied by the occurrence of new objects.
Besides, it is worth noting that LaMD only supports image-conditional video generation tasks at present. To expand its capabilities and address broader applications such as text-to-video generation, combining the LaMD with advanced image generation models is considered to generate appearance and motion simultaneously.
In addition, LaMD is currently evaluated on close-domain datasets. However, experiments demonstrating enhanced capabilities with a smaller model size and effectiveness across various motion patterns suggest that our method has potential for open-domain video generation. We view this as an exciting avenue for future work.

\noindent
\textbf{Acknowledgments}
This work was supported in part by the National Natural Science Foundation of China under contract No. 62036005. The numerical calculations in this paper have been done on the supercomputing system in the Supercomputing Center of Wuhan University.

\end{sloppypar}

\small{
    \bibliographystyle{spbasic}
    \bibliography{sn-bibliography}
}

\end{document}